\documentclass[10pt,twocolumn,letterpaper]{article}

\usepackage{cvpr}              

\usepackage{colortbl}
\usepackage{xcolor}
\usepackage[accsupp]{axessibility}
\definecolor{cvprblue}{rgb}{0.21,0.49,0.74}
\usepackage[pagebackref,breaklinks,colorlinks,allcolors=cvprblue]{hyperref}

\def\paperID{41578} 
\def\confName{CVPR}
\def\confYear{2026}

\title{Masking Matters: Unlocking the Spatial Reasoning Capabilities of LLMs \\ for 3D Scene-Language Understanding}

\author{Yerim Jeon, Miso Lee, WonJun Moon, Jae-Pil Heo\thanks{Corresponding author}\\
Sungkyunkwan University\\
{\tt\small \{1357j, dlalth557, wjun0830, jaepilheo\}@skku.edu}
}

\usepackage[utf8]{inputenc} 
\usepackage[T1]{fontenc}    
\usepackage{hyperref}       
\usepackage{url}            
\usepackage{booktabs}       
\usepackage{amsfonts}       
\usepackage{nicefrac}       
\usepackage{microtype}      
\usepackage{xcolor}         
\usepackage{kotex}         
\usepackage{tabularx}
\usepackage{booktabs}
\usepackage{multirow}
\newcolumntype{Y}{>{\centering\arraybackslash}X}
\usepackage{graphicx}
\usepackage{amsmath}
\usepackage{amsthm}
\usepackage[many]{tcolorbox}
\usepackage{algorithm}
\usepackage{algorithmicx}
\usepackage{algpseudocode}
\usepackage{mathtools,amssymb,nccmath}
\algrenewcommand{\algorithmicindent}{0.5em}
\usepackage{amsfonts}       
\usepackage[table,xcdraw]{xcolor}
\definecolor{Light}{rgb}{0.99, 0.92, 0.95}
\usepackage{pifont}
\usepackage{color} 
\usepackage{soul} 
\usepackage{tcolorbox}
\usepackage{xcolor}
\usepackage{colortbl}

\newcommand{\xmark}{\ding{55}}%

\begin{document}
\maketitle
\begin{abstract}
Recent advances in 3D scene-language understanding have leveraged Large Language Models (LLMs) for 3D reasoning by transferring their general reasoning ability to 3D multi-modal contexts. However, existing methods typically adopt standard decoders from language modeling, which rely on a causal attention mask. This design introduces two fundamental conflicts in 3D scene understanding: sequential bias among order-agnostic 3D objects and restricted object-instruction attention, hindering task-specific reasoning. To overcome these limitations, we propose 3D Spatial Language Instruction Mask (3D-SLIM), an effective masking strategy that replaces the causal mask with an adaptive attention mask tailored to the spatial structure of 3D scenes. Our 3D-SLIM introduces two key components: a Geometry-adaptive Mask that constrains attention based on spatial density rather than token order, and an Instruction-aware Mask that enables object tokens to directly access instruction context. This design allows the model to process objects based on their spatial relationships while being guided by the user's task. 3D-SLIM is simple, requires no architectural modifications, and adds no extra parameters, yet it yields substantial performance improvements across diverse 3D scene-language tasks. Extensive experiments across multiple benchmarks and LLM baselines validate its effectiveness and underscore the critical role of decoder design in 3D multi-modal reasoning. The code is available at \url{https://github.com/Jyerim/3D-SLIM}.
\end{abstract}    
\section{Introduction}
\label{sec:intro}

\begin{figure}[t]
    \centering
    \includegraphics[width=1.0\linewidth]{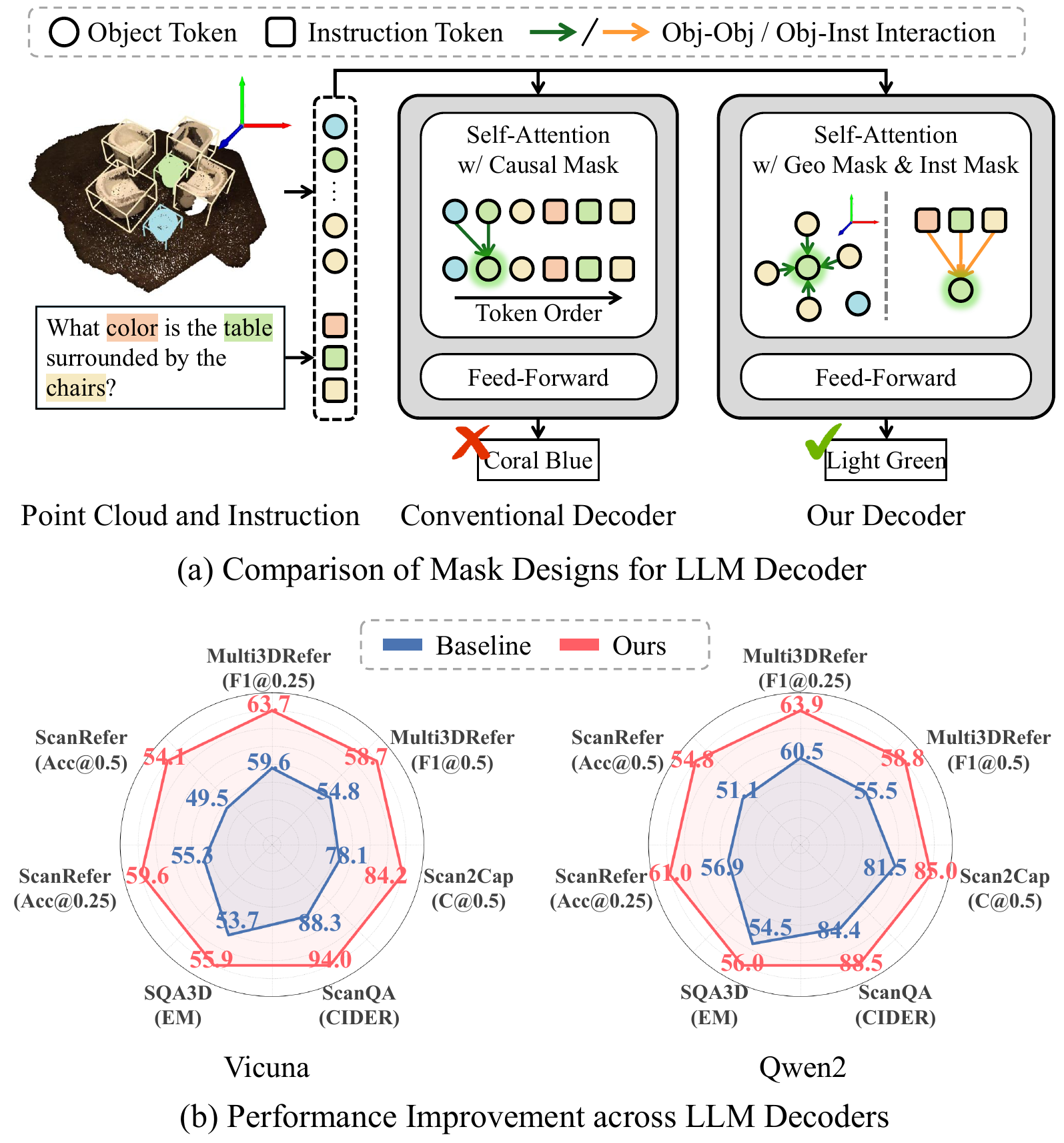}
    \caption{\textbf{Comparison of Mask Designs and Impact on Performance.}
    (a) Unlike conventional causal masking that reinforces the input order, 3D-SLIM introduces Geo Mask and Inst Mask to capture spatial structures and encode task-oriented object representations for enhanced 3D reasoning.
    (b) When integrated with various LLM decoders, 3D-SLIM consistently improves performance across multiple benchmarks.}
    \label{fig:intro}
\end{figure}
3D scene-language understanding aims to jointly interpret 3D environments and natural language, serving as a fundamental basis for multi-modal reasoning in applications such as robotic navigation and embodied agents.
Traditionally, 3D scene-language tasks have relied on specialized, task-specific models~(e.g., grounding, captioning, and question-answering).
With the emergence of Large Language Models~(LLMs), research has moved toward unified frameworks capable of handling multiple tasks within a single model.

Recently, Chat-Scene~\cite{chatscene} introduced an Object-Centric 3D LLM framework that decomposes an input 3D scene into a set of object proposals using a pretrained detector~\cite{mask3d}.
Each object is represented by an identifier token and its instance-level 3D and 2D features, enabling object-centric reasoning via LLMs~\cite{vicuna,llama3}.
Following this paradigm, Inst3D-LMM~\cite{inst3dlmm} incorporates 2D-3D feature interactions into object-level representations, and 3DGraphLLM~\cite{3dgraphllm} further models semantic relationships among objects.
These works have achieved promising performance with compact yet expressive scene representations.

However, this significant progress has been largely confined to input representation, leaving the decoder architecture relatively underexplored.
Current approaches borrow decoders directly from language modeling~\cite{vicuna,llama3}, neglecting their inherent mismatch with 3D data.
We identify two fundamental conflicts arising from the standard causal mask. 
First, it imposes a sequential dependency on object tokens, which contradicts the order-agnostic nature of 3D scenes. 
This forces the model to learn spurious order-dependent correlations, although 3D scenes are intrinsically organized by spatial relationships rather than input order.
Second, the causal mask limits essential interaction between object tokens and the instruction tokens, compelling the model to process the entire 3D scene before integrating the user's instruction, leading to inefficient reasoning pathways.

To overcome these limitations, we propose 3D Spatial Language Instruction Mask~(3D-SLIM), a masking strategy that replaces the causal mask with an adaptive attention mask tailored to the spatial structure of 3D scenes, as illustrated in \cref{fig:intro}.
We first introduce a Geometry-adaptive Mask~(Geo Mask), which models local object relationships by adaptively determining the attention scope based on each object's spatial density.
This mask dynamically adjusts the attention range, permitting more accessible neighbors in object-dense areas to capture rich local context while restricting them in sparse areas to prevent attending to irrelevant, distant objects.
Subsequently, we introduce an Instruction-aware Mask~(Inst Mask) that enables direct attention from object tokens to the instruction tokens. 
This modification allows the model to leverage the instruction context during processing each object, thereby decoding the scene into an effective, task-adapted representation.

3D-SLIM is a simple yet highly effective masking strategy that directly substitutes the standard causal mask.
This design enables seamless integration into various LLM decoders without any architectural modifications or additional parameters.
The generality and effectiveness of our method have been validated through extensive experiments across diverse baselines on multiple 3D scene-language tasks, as shown in \cref{fig:intro}.
We believe this work provides new insights into the underlying importance of decoder design for advancing 3D scene-language understanding.

Our key contribution can be summarized as follows:
\begin{itemize}
    \setlength\itemsep{5pt}
    \item We identify two major issues in the conventional LLM decoders: spurious order-dependent correlations and limited interaction between instruction and object tokens.
    \item We propose a 3D Spatial Language Instruction Mask~(3D-SLIM), an efficient and easily integrable solution to address both issues.
    \item We demonstrate the effectiveness and high applicability of 3D-SLIM across various baselines and multiple 3D scene-language tasks.
\end{itemize} 

\section{Related Work}
\paragraph{3D Scene-Language Understanding.}
3D scene understanding has emerged as a key computer vision task, enabling models to interpret complex 3D environments and solve diverse spatial reasoning problems based on user instructions.
This field involves fundamental 3D vision-language tasks, such as 3D visual grounding~\cite{scanrefer, multi3drefer}, 3D dense captioning~\cite{scan2cap}, and 3D question answering~\cite{scanqa, sqa3d}, which require a comprehensive understanding of spatial structures and semantic relationships within 3D scenes.
Early approaches~\cite{eda,vote2cap++, dspnet} tackled these challenges with task-specific models, while subsequent studies~\cite{3djcg, 3dvista, 3dvlp} proposed unified frameworks that jointly handle multiple tasks.
However, these approaches still rely on task-specific heads, limiting their flexibility for more generalized user instructions.

\paragraph{Large Language Models for 3D Scene Understanding.}
Motivated by the powerful generalization capabilities of Large Language Models~(LLMs)~\cite{vicuna,llama3,qwen2}, numerous studies have investigated extending LLMs to 3D scene understanding.
The key challenge in this extension is to construct 3D scene representations that are well aligned with the input modality of LLMs.
To address this challenge, recent studies have evolved into three main directions based on how 3D scenes are represented.
One line of research focuses on point-based representations~\cite{3dllm, scenellm,llava3d}, which project multi-view image features onto 3D point or voxel spaces to construct dense geometric representations.
Another direction explores video-based representations~\cite{video3dllm, gpt4scene, ross3d}, treating multi-view images as video sequences and injecting 3D information into video LLMs~\cite{qwen2-vl,llavavideo} to leverage their strong multi-modal reasoning capabilities.
In contrast, object-based approaches~\cite{chat3d, chat3dv2, grounded3dllm, leo, inst3dlmm, robin3d, chatscene, 3dgraphllm} abstract 3D scenes into object-centric representations and assign identifiers for each object to facilitate direct reference and grounding.
While all these approaches primarily focus on input-level organization, our work shifts the focus toward decoder design, tailoring it to jointly process complex 3D scene representations and language instructions.
\section{Preliminary and Motivation}
\label{sec:preliminary}
\begin{figure*}[t]
    \centering
    \includegraphics[width=1.0\linewidth]{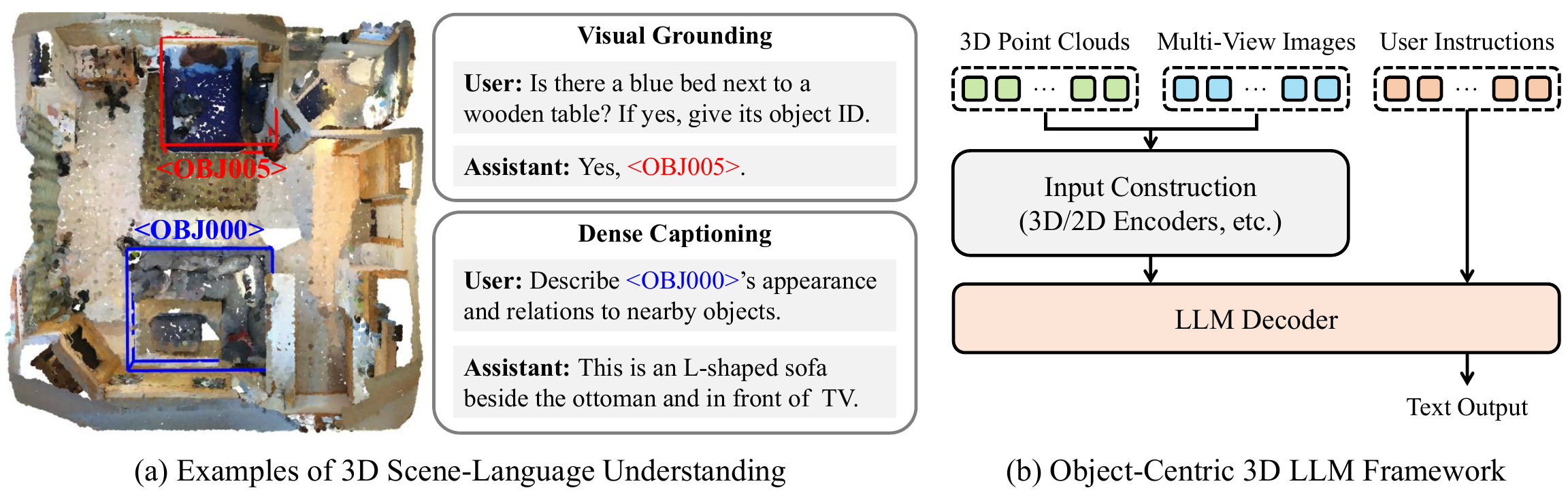}
    \caption{
    \textbf{Overview of 3D Scene-Language Understanding and an Object-Centric Framework.} 
    (a) Examples of 3D reasoning tasks, including visual grounding and dense captioning. 
    Object identifiers~(e.g., <OBJ000>) are utilized to reference and ground target objects during the conversation. 
    (b) The object-centric 3D LLM consists of an input construction module and an LLM decoder. 
    Unlike prior works focusing on input construction, we emphasize the decoder to enhance 3D reasoning.
    }
    \label{fig:preliminary}
\end{figure*}
\paragraph{Task Definition.}
As depicted in \cref{fig:preliminary}, 3D scene-language understanding integrates 3D perception with natural-language reasoning to interpret complex 3D environments.
Given a 3D scene and a user instruction, the model learns to identify referred objects~\cite{scanrefer, multi3drefer}, describes specified targets~\cite{scan2cap}, or answers questions~\cite{scanqa, sqa3d}.
Such tasks require understanding objects’ visual appearance and their spatial and semantic relationships within the scene.

\paragraph{Object-Centric Scene Understanding.}
Our work builds upon object-centric 3D LLM frameworks~\cite{chatscene, inst3dlmm, 3dgraphllm}, which provide a representative pipeline for 3D scene-language understanding. 
This pipeline consists of two stages, as illustrated in \cref{fig:preliminary}:

\noindent\textbf{(1) Input Construction} forms object-centric representations by decomposing the 3D scene into object proposals using a pretrained detector~\cite{mask3d} and encoding each object with geometric, visual and relational cues.
Object identifiers are then assigned to these objects and incorporated into the language model’s vocabulary to reason with explicit object references.
The resulting set of object tokens~(each containing an identifier and its features) is arranged together with system and instruction tokens, forming a complete multi-modal sequence for the LLM decoder, which is structured in the specific order of \texttt{[system, objects, instruction]}.

\noindent\textbf{(2) LLM Decoder} serves as the core reasoning module, processing the multi-modal input sequences to generate task-specific outputs.
Following the standard design of decoder-only architectures~\cite{llama, llama2, llama3, vicuna, qwen2}, the decoder operates in an auto-regressive manner, where each token attends only to preceding ones by applying a causal mask in self-attention.
Given that $Q,K,V\in \mathbb{R}^{n\times{d}}$ are the query, key, and value matrices derived from the input sequence $X$, the attention weights are computed as 
$A=\text{softmax}\!\left(\frac{QK^T}{\sqrt{d}}+M\right)$, where $n$ and $d$ denotes the sequence length and feature dimension, respectively.
The mask $M$ is defined such that its elements $M_{pq} = -\infty$ for $p < q$, which ensures that the resulting attention weight matrix $A$ is lower triangular. 

\paragraph{Motivation.}
Although extensive efforts have focused on organizing input sequences, the decoder design for 3D scene understanding has remained underexplored.
Specifically, the causal masking mechanism in conventional decoders fundamentally conflicts with the order-agnostic nature of 3D scenes. 
While crucial for text, where token order encodes syntax and semantics, the causal mask is misaligned with 3D object tokens that have no inherent sequential order.
This property forces an artificial sequential interaction, causing the decoder to prioritize the arbitrary input order over the scene's underlying geometric relationships.
Moreover, the causal constraint blocks interaction between object and instruction tokens, hindering the integration of linguistic cues into scene interpretation.
This disconnection forces the model to interpret scenes independently of the instruction, leading to inefficient cross-modal reasoning. 
These limitations highlight the need for a decoder tailored to the spatial and multi-modal characteristics of 3D scene-language understanding, as elaborated in \cref{sec:method}.
\section{Method}
\label{sec:method}
\begin{figure*}[t]
    \centering
    \includegraphics[width=1.0\linewidth]{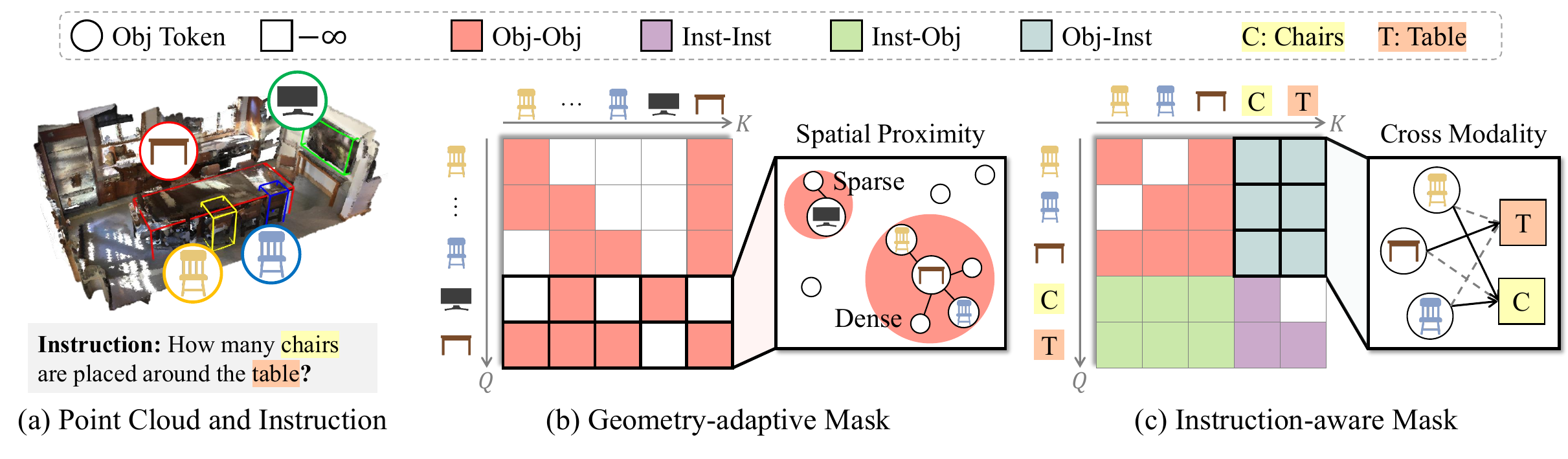}
    \caption{
    \textbf{Overview of 3D Spatial Language Instruction Mask~(3D-SLIM).}
    (a) Illustration of the input 3D point cloud and user instruction, which are subsequently encoded as object and instruction tokens.
    For simplicity, only a small subset of object and instruction tokens is depicted.
    (b) The Geometry-adaptive Mask dynamically modulates object-object attention based on spatial proximity and local density; objects in denser regions attend to an expanded set of neighbors, whereas those in sparse areas focus on fewer neighbors.
    (c) The Instruction-aware Mask enables object-instruction attention, which enhances scene understanding by focusing on instruction-relevant content~(e.g., "chairs", "table").}
    \label{fig:overview}
\end{figure*}
To address the aforementioned challenges, we propose 3D Spatial Language Instruction Mask~(3D-SLIM) to tailor LLM decoders for 3D scene understanding.
As illustrated in \cref{fig:overview}, 3D-SLIM is composed of two specialized masking schemes designed to govern object-object and object-instruction interactions.
First, we introduce the Geometry-adaptive Mask~(Geo Mask) to promote interaction among spatially proximate objects. 
By considering both distance and local object density, this mask supports a geometrically grounded understanding of the 3D scene's internal structure.
Subsequently, we apply the Instruction-aware Mask~(Inst Mask) to facilitate direct interaction between instruction and object tokens. 
Both components seamlessly integrate into existing decoders, requiring no architectural modifications.

\paragraph{Geometry-adaptive Mask.}

The causal mask imposes sequential dependencies among object tokens, which contradicts the order-agnostic nature of 3D scenes.
A na\"ive solution is to allow full attention across all object tokens; however, this relaxation of sequential constraints disregards the inherent local structure of 3D scenes by treating all object relationships uniformly.
In contrast, humans tend to interpret 3D scenes by grouping nearby objects according to spatial proximity~\cite{gestalt}, thereby grasping complex 3D scenes through the formation of locally coherent structures.
We suggest that incorporating this proximity bias into the model can enable more effective reasoning about the spatial structure of 3D scenes.
Motivated by this idea, we design a Geo Mask that adaptively adjusts the attention scope based on local object density. 

Concretely, Geo Mask specifies a local neighborhood to model spatially proximate relationships among objects.
However, pre-defined thresholds~(e.g., number of objects and fixed distance) are insufficient for identifying local neighbors, as
3D scenes have non-uniform spatial object density, containing both densely clustered and sparsely distributed regions.
To handle this variation, Geo Mask begins by estimating the local density around each object, from which an adaptive number of neighboring objects is inferred.
Specifically, the local density of $i$-th object is computed as:
\begin{equation}
\begin{aligned}
\rho_{i}&=\sqrt{3}-\frac{1}{N-1}\sum_{j:j \neq i}d_{ij}, \ \ 
\tilde{\rho}_{i}= \frac{\rho_{i} - \rho_{\min}}{\rho_{\max} - \rho_{\min}}, \\
\end{aligned}
\label{eq:local_density}
\end{equation}
where $i,j \in [1,N]$ denote the object indices within the object token set~$\mathcal{O}$, $N$ is the total number of objects, and
$d_{ij} = \|{\mathbf{c}}_i - \mathbf{c}_j\|_2$ corresponds to the L2 distance between object centers.
$\rho_{\min}$ and $\rho_{\max}$ indicate the minimum and maximum local density values across all objects, and $\tilde{\rho}$ is the corresponding min-max normalized density that supports relative local density comparison within the scene.
Thus, objects with smaller average distances to others exhibit higher local object density.

Subsequently, we compute the adaptive number of local neighbors and the corresponding attentive neighborhood for each object as follows:
\begin{equation}
\begin{aligned}
k_i &= \operatorname{round}\big((k_{\max} - k_{\min})\cdot \tilde{\rho_i} + k_{\min}\big), \\
\Omega_i &= \operatorname{TopK}(\{-d_{ij} \mid j \in \{1, \dots , N\} \setminus \{i\}\}, k_i), \\
\end{aligned}
\label{eq:adaptive_local_neighbors}
\end{equation}
where $k_i \in [k_{\min}, k_{\max}]$ defines the adaptive attention range for $i$-th object,
$\operatorname{round}(\cdot)$ denotes rounding to the nearest integer, 
and $\operatorname{TopK}(D, k)$ is the operation that selects the indices of $k$ largest elements in $D$, respectively.
Using the neighbor set $\Omega_i$, the attention mask that determines object-to-object interactions is finally defined as:
\begin{equation}
    \begin{aligned}
        &M_{ij}^{\mathcal{O}} =\begin{cases} 
            0, & j \in \Omega_i \text{ or } j=i,\\
            -\infty, & \text{otherwise.} 
            \end{cases}
    \end{aligned}
\label{eq:nnmask}
\end{equation}
It captures the structure of non-uniform 3D scenes by permitting broader attention in dense regions while limiting it in sparse ones.
This behavior can be interpreted as a geometry-aware scene graph that dynamically adjusts local connectivity based on spatial density.

\paragraph{Instruction-aware Mask.}
Humans naturally interpret visual scenes through linguistic cues.
In other words, when given a question or instruction, they focus on regions relevant to the described context.
From this perspective, we claim that instruction-aware object encoding is crucial for effective multi-modal reasoning in LLMs, yet this process is hindered by the causal mask.

To address this issue, the Inst Mask is introduced, allowing attention from object tokens to instruction tokens.
Specifically, the Inst Mask overrides the object-instruction block of the attention mask $M$ by replacing the $-\infty$ entries with zeros. 
The Inst Mask is defined as:
\begin{equation}
    \begin{aligned}
        &M_{ij}^{\mathcal{I}} =\begin{cases} 
            0, & i \in \mathcal{O} \text{ and } j \in \mathcal{I},\\
            M_{ij}, & \text{otherwise,} 
            \end{cases}
    \end{aligned}
\label{eq:nnmask}
\end{equation}
where $\mathcal{O}$ and $\mathcal{I}$ denotes the object and instruction token sets, respectively. 
This simple adjustment restores the instruction-to-object attention pathway, aligning object representations with the linguistic context for more effective cross-modal understanding.

\paragraph{Training Objective.}
Following Chat-Scene~\cite{chatscene}, we unify all 3D scene-language tasks into a single input-output format, facilitating the model's ability to handle diverse task formulations within one framework.
Under this unified prompt, the model is trained solely with the cross-entropy loss that supervises text generation conditioned on the multi-modal inputs.
Given the input sequence $X$~(consisting of the system message, scene representation, and user instruction), the training objective minimizes the negative log-likelihood of the target sequence $Y$:
\begin{equation}
    \begin{aligned}
        &\mathcal{L} = - \sum_{l=1}^{m}\text{log}P(Y_l|Y_{[1,\ldots, l-1]}, X),
    \end{aligned}
\label{eq:objective}
\end{equation}
where $m$ denotes the number of tokens in the target sequence, and $Y_{1,\ldots,l-1}$ represents the sequence of the previous $l-1$ tokens in the response.

\definecolor{sectiongray}{gray}{0.93}
\definecolor{linegray}{gray}{0.7}

\begin{table*}[t]
\centering
\resizebox{0.95\linewidth}{!}{
\begin{tabular}{l|c|cccc|cc|cccccc}
\toprule
  \multirow{3}{*}{\textbf{Method}} &
  \multirow{3}{*}{\textbf{LLM}} &
  \multicolumn{4}{c|}{\textbf{3D Visual Grounding}} &
  \multicolumn{2}{c|}{\textbf{3D Captioning}} &
  \multicolumn{6}{c}{\textbf{3D Question-Answering}} \\[0.6ex]

  &&
  \multicolumn{2}{c}{ScanRefer$_{\text{val}}$} &
  \multicolumn{2}{c|}{Multi3DRefer$_{\text{val}}$} &
  \multicolumn{2}{c|}{Scan2Cap$_{\text{val}}$} &
  \multicolumn{2}{c}{ScanQA$_{\text{val}}$} &
  \multicolumn{2}{c}{SQA3D$_{\text{val}}$} &
  \multicolumn{2}{c}{SQA3D$_{\text{test}}$} \\
 & & Acc@0.25 & Acc@0.5 & F1@0.25 & F1@0.5 & C@0.5 & B-4@0.5 & C & B-4 & EM   & EM-R & EM   & EM-R \\ \midrule
\rowcolor{sectiongray}
\multicolumn{14}{l}{\textit{Expert Models}} \\
\arrayrulecolor{linegray}\midrule
 ScanRefer~\cite{scanrefer} & \xmark & 37.3 & 24.3 & - & - & - & - & - & - & - & - & - & - \\
 ScanQA~\cite{scanqa}  & \xmark & - & - & - & - & - & - & 64.9  & 10.1 & -    & - & - & -         \\
 Scan2Cap~\cite{scan2cap} & \xmark & - & - & - & - & 81.3 & 43.4 & -  & - & -   & - & - & -         \\
 SQA3D~\cite{sqa3d} & \xmark & - & - & - & - & - & - & -  & - & -   & - & 46.6 & -         \\
 3DJCG~\cite{3djcg}  & \xmark & 49.6 & 37.3 & - & - & 49.5 & 31.0 & - & - & - & - & - & -       \\
 3D-VLP~\cite{3dvlp} & \xmark & 51.4 & 39.5 & - & - & 54.9 & 32.3 & 67.0 & 11.1 & - & -  & - & -        \\
 M3DRef-CLIP~\cite{multi3drefer}  & \xmark & 51.9 & 44.7 & 42.8 & 38.4 & - & - & - & - & - & -  & - & -  \\
 3D-VisTA~\cite{3dvista}  & \xmark & 50.6 & 45.5 & - & - & 66.9 & 34.0 & 72.9  & 13.1 & - & - & 48.5 & -          \\
 PQ3D~\cite{PQ3D} & \xmark & - & 51.2 & - & 50.1 & 80.3 & 36.0 & 87.8 & - & - & - & 47.1 & - \\
\arrayrulecolor{black}\midrule
\rowcolor{sectiongray}
\multicolumn{14}{l}{\textit{Point-Based Representation}} \\
\arrayrulecolor{linegray}\midrule
 3D-LLM~\cite{3dllm} & BLIP2-flant5 & 30.3 & - & - & - & - & - & 69.4  & 12.0   & -    & -  & - & -        \\
 Scene-LLM~\cite{scenellm} & Llama-2-7B & - & - & - & - & - & - & 80.0  & 12.0   & - & - & 54.2 & -          \\
 LLaVA-3D~\cite{llava3d} & LLaVA-7B-v1.5 & 54.1 & 42.4 & - & - & 79.2 & 41.1 & 91.7 & - & - & - & 55.6 & 57.6 \\
 
\arrayrulecolor{black}\midrule
\rowcolor{sectiongray}
\multicolumn{14}{l}{\textit{Video-Based Representation}} \\
\arrayrulecolor{linegray}\midrule
 Qwen2-VL-7B$^{\dagger}$~\cite{qwen2-vl} & Qwen2-VL-7B & - & - & - & - & - & - & 53.9 & 3.0 & - & - & 40.7 & 46.7 \\
 LLaVA-Video-7B$^{\dagger}$~\cite{llavavideo} & LLaVA-Video-7B-Qwen2 & - & - & - & - & - & - & 88.7 & 3.1 & - & - & 48.5 & - \\ 
 GPT4Scene-HDM~\cite{gpt4scene} & Qwen2-VL-7B & 62.6 & 57.0 & 64.5 & 59.8 & 86.3 & 40.6 & 96.3 & 15.5 & - & - & 59.4 & 62.4 \\
 Video-3D LLM~\cite{video3dllm} & LLaVA-Video-7B-Qwen2 & 58.1 & 51.7 & 58.0 & 52.7 & 83.8 & 41.3 & 102.1 & 16.4 & - & -  & 58.6 & - \\
 Ross3D~\cite{ross3d} & LLaVA-Video-7B-Qwen2 & 61.1 & 54.5 & 59.6 & 54.3 &81.3 & \textbf{43.4} & \textbf{107.0} & \textbf{17.9} & - & -  & \textbf{63.0}  & \textbf{65.7} \\

\arrayrulecolor{black}\midrule
\rowcolor{sectiongray}
\multicolumn{14}{l}{\textit{Object-Based Representation}} \\
\arrayrulecolor{linegray}\midrule
 Chat-3D~\cite{chat3d} & Vicuna-7B-v0 & - & - & - & - & - & - & 53.2  & 6.4    & -    & -   & - & -        \\
 Chat-3D v2~\cite{chat3dv2} & Vicuna-7B-v1.5 & 42.5 & 38.4 & 45.1 & 41.6 & 63.9 & 31.8 & 87.6  & 14.0  & - & -  & 54.7  & -          \\
 Grounded 3D-LLM~\cite{grounded3dllm} & Tiny-Vicuna-1B & 47.9 & 44.1 & 45.2 & 40.6 & 70.6 & 35.5 & 72.7 & 13.4 & - & - & - & - \\
 LEO~\cite{leo} & Vicuna-7B-v1.1 & - & - & - & - & 68.4 & 36.9 & 80.0  & 11.5   & - & -  & 50.0   & 52.4       \\
 Inst3D-LMM~\cite{inst3dlmm} & Vicuna-7B-v1.5 & 57.8 & 51.6 & 58.3 & 53.5 & 79.7 & 38.3 & 88.6 & 14.9 & - & -  & - & - \\
 Robin3D~\cite{robin3d} & Vicuna-7B-v1.5 & 60.8 & 55.1 & 64.9 & 59.7 & \textbf{87.2} & 38.4 & - & - & 56.0 & 58.6  & 56.9 & 59.8 \\
 \midrule
 Chat-Scene~\cite{chatscene} & Vicuna-7B-v1.5 & 55.5 & 50.2 & 57.1 & 52.4 & 77.1 & 36.3 & 87.7 & 14.3 & 53.2 & 56.1 & 54.6 & 57.5     \\ 
 \rowcolor{Light}Chat-Scene + Ours & Vicuna-7B-v1.5 & 59.6 & 54.1 & 63.7 & 58.7 & 84.2 & 38.0 & 94.0 & 15.2 & 55.9 & 58.9 & 55.5 & 58.2    \\ 
 3DGraphLLM~\cite{3dgraphllm} & Llama3-8B-Instruct & 62.4 & 56.6 & 64.7 & 59.9 & 81.0 & 36.5 & 88.8 & 15.9 & 55.9 & - & - & -  \\
 \rowcolor{Light}3DGraphLLM + Ours & Llama3-8B-Instruct & \textbf{64.1} & \textbf{57.7} & \textbf{67.3} & \textbf{62.0} & 82.2 & 37.3 & 88.2 & 15.8 & \textbf{56.8} & \textbf{59.7} & 56.1 & 59.1  \\
 \arrayrulecolor{black}\bottomrule 
 \end{tabular}}
\caption{\textbf{Performance comparison on 3D scene-language understanding benchmarks.} The table shows the evaluation results of Expert models and three types of Generalist models: Point-based, Video-based, and Object-based models. "EM" stands for top-1 exact match and "EM-R" means the refined exact match following~\cite{leo}. $\dagger$ indicates zero-shot performance and $-$ denotes unavailable results.} \label{tab:performance_comparison}
\end{table*}
\section{Experiments}
\subsection{Datasets and Metrics}
\paragraph{Datasets.}
We conduct experiments on five widely used benchmarks for 3D scene-language understanding.
For visual grounding, we evaluate on ScanRefer~\cite{scanrefer} and Multi3DRefer~\cite{multi3drefer}, which aim to identify single-object and multi-object targets, respectively.
For dense captioning, we use Scan2Cap~\cite{scan2cap}, which assesses how well the model describe each object's appearance and its spatial relations with surrounding objects.
For question answering, we adopt ScanQA~\cite{scanqa} and SQA3D~\cite{sqa3d}, which target free-form question-answering and situated reasoning, respectively.
All benchmarks are built upon ScanNet~\cite{scannet}, a large-scale indoor dataset providing 3D point clouds, RGB-D sequences, and instance-level segmentation annotations.
It contains 1513 scenes in total, with 1201 used for training and 312 for validation.
Following prior works~\cite{chatscene, inst3dlmm, 3dgraphllm}, we report results on the validation sets for all benchmarks, while for SQA3D we additionally follow the standard practice of reporting test set performance.

\paragraph{Metrics.}
We follow the standard evaluation metrics used in previous works~\cite{chatscene,inst3dlmm,3dgraphllm} for each benchmark. 
For ScanRefer, we report Acc@0.25 and Acc@0.5, which indicate the percentage of predictions whose 3D IoU with the ground-truth bounding box exceeds 0.25 and 0.5, respectively. 
For Multi3DRefer, we use F1@0.25 and F1@0.5, combining precision and recall under the same IoU thresholds. 
For Scan2Cap, evaluation is based on C@0.5~(CIDEr@0.5) and B-4@0.5~(BLEU-4@0.5), which measure the similarity between generated and reference captions conditioned on the 3D IoU threshold of 0.5. 
For ScanQA, we report C~(CIDEr)~\cite{cider} and B-4~(BLEU-4)~\cite{blue}, assessing the linguistic similarity between predicted and reference answers. 
Finally, for SQA3D, we adopt EM~(Exact Match)~\cite{sqa3d} and EM-R~(Refined Exact Match)~\cite{leo}, where EM requires exact matching with the ground-truth answer, and EM-R denotes a refined protocol for answer matching.

\subsection{Implementation Details}
We assess the effectiveness and adaptability of 3D-SLIM by integrating it into two representative frameworks, Chat-Scene~\cite{chatscene} and 3DGraphLLM~\cite{3dgraphllm}.
For hyperparameters, we adopt their original configurations.
All models are fine-tuned with LoRA using the AdamW optimizer~(weight decay 0.02), and Non-Maximum Suppression~(NMS) with a mask IoU threshold of 0.9 is applied.
We set the batch size to 32 for Chat-Scene and 8 for 3DGraphLLM, with learning rates of $5e^{-6}$ and $2e^{-5}$, respectively.
For Geo Mask, we empirically set the lower and upper bounds of the attentive neighborhood to $k_{\min}=2$ and $k_{\max}=10$, respectively.
All experiments are conducted on 2 NVIDIA RTX Pro6000 GPUs.

\begin{table*}[t]
\label{tab:ablation study}
\centering
\small
\setlength{\tabcolsep}{4pt}
\renewcommand{\arraystretch}{0.9}
\resizebox{0.95\linewidth}{!}{
\begin{tabular}{l|cccccccccc}
\toprule
    \multirow{2}{*}{Method} &
    \multicolumn{2}{c}{ScanRefer$_{\text{val}}$} &
    \multicolumn{2}{c}{Multi3DRefer$_{\text{val}}$} &
    \multicolumn{2}{c}{Scan2Cap$_{\text{val}}$} &
    \multicolumn{2}{c}{ScanQA$_{\text{val}}$} &
    \multicolumn{2}{c}{SQA3D$_{\text{val}}$} \\
      &  Acc@0.25 & Acc@0.5 & F1@0.25 & F1@0.5 & C@0.5 & B-4@0.5 & C & B-4 & EM   & EM-R \\ 
\midrule
    Vicuna-7B-v1.5~\cite{vicuna} & 55.3 & 49.5 & 59.6 & 54.8 & 78.1 & 35.6 & 88.3 & 14.1 & 53.7 & 56.4 \\
    \rowcolor{Light} Vicuna-7B-v1.5 + Ours & \textbf{59.6} & \textbf{54.1} & \textbf{63.7} & \textbf{58.7} & \textbf{84.2} & \textbf{38.0} & \textbf{94.0} & \textbf{15.2} & \textbf{55.9} & \textbf{58.9} \\
    Llama3-8B-Instruct~\cite{llama3} & 59.8 & 53.9 & 61.9 & 57.2 & 83.9 & 38.1 & 82.5 & 14.1 & 56.1 & 59.1 \\
    \rowcolor{Light} Llama3-8B-Instruct + Ours & \textbf{61.8} & \textbf{55.7} & \textbf{64.3} & \textbf{59.2} & \textbf{85.1} & \textbf{38.8} & \textbf{85.1} & \textbf{15.6} & \textbf{56.3} & \textbf{59.2} \\
    Qwen2-7B-Instruct~\cite{qwen2} & 56.9 & 51.1 & 60.5 & 55.5 & 81.5 & 37.7 & 84.4 & 14.9 & 54.5 & 57.4 \\
    \rowcolor{Light} Qwen2-7B-Instruct + Ours & \textbf{61.0} & \textbf{54.8} & \textbf{63.9} & \textbf{58.8} & \textbf{85.0} & \textbf{38.9} & \textbf{88.5} & \textbf{15.9} & \textbf{56.0} & \textbf{59.2} \\
    Qwen3-8B-Instruct~\cite{qwen3} & 57.5 & 51.5 & 61.3 & 56.5 & 78.8 & 36.7 & 83.0 & 14.6 & 54.7 & 58.3 \\
    \rowcolor{Light} Qwen3-8B-Instruct + Ours & \textbf{61.7} & \textbf{55.8} & \textbf{64.5} & \textbf{59.6} & \textbf{83.6} & \textbf{38.7} & \textbf{84.8} & \textbf{15.0} & \textbf{55.7} & \textbf{59.2} \\
\bottomrule
\end{tabular}
}
\caption{\textbf{Performance across LLM Decoders.} The table shows the performance of 3D-SLIM with different LLM decoders under the Chat-Scene framework.} 
\vspace{-0.5em}
\label{tab:various llm}
\end{table*}

\begin{table*}[t]
\label{tab:ablation study}
\centering
\small
\setlength{\tabcolsep}{4pt}
\renewcommand{\arraystretch}{0.9}
\begin{tabular}{ll|cccccccccc}
\toprule
    \multirow{2}{*}{Row} &
    \multirow{2}{*}{Method} &
    \multicolumn{2}{c}{ScanRefer$_{\text{val}}$} &
    \multicolumn{2}{c}{Multi3DRefer$_{\text{val}}$} &
    \multicolumn{2}{c}{Scan2Cap$_{\text{val}}$} &
    \multicolumn{2}{c}{ScanQA$_{\text{val}}$} &
    \multicolumn{2}{c}{SQA3D$_{\text{val}}$} \\
    & &  Acc@0.25 & Acc@0.5 & F1@0.25 & F1@0.5 & C@0.5 & B-4@0.5 & C & B-4 & EM   & EM-R \\ 
\midrule
    A0 & Causal Mask & 55.3 & 49.5 & 59.6 & 54.8 & 78.1 & 35.6 & 88.3 & 14.1 & 53.7 & 56.4 \\
    \midrule
    B0 & Full Mask$^{\dagger}$ & 55.9 & 50.3 & 60.1 & 55.3 & 76.5 & 35.3 & 87.9 & 15.1 & 53.2 & 55.9 \\
    \midrule
    C0 & Full Mask & 56.2 & 50.5 & 61.2 & 56.0 & 78.4 & 36.0 & 90.9 & 14.8 & 54.8 & 57.6 \\
    C1 & Diagonal Mask & 56.4 & 50.6 & 60.5 & 55.6 & 78.6 & 36.4 & 92.9 & 14.9 & 53.9 & 56.7 \\
    \midrule
    D0 & Fixed-N Mask & 57.5 & 51.7 & 61.6 & 56.6 & 81.9 & 36.7 & 91.6 & 15.0 & 55.3 & 58.0 \\
    D1 & Geo Mask~(Ours) & \textbf{58.6} & \textbf{53.1} &\textbf{62.0} & \textbf{57.3} & \textbf{82.4} & \textbf{37.0} & \textbf{94.2} & \textbf{15.0} & \textbf{55.9} & \textbf{58.6} \\
\bottomrule
\end{tabular}
\caption{\textbf{Ablation study on Decoder Masking Strategies.} The table presents the results of different masking strategies under the Chat-Scene framework. A0~(Causal Mask) is the standard causal mask adopted in existing LLM decoders. B0~(Full Mask$^{\dagger}$) is the full attention applied over the entire mask. C0~(Full Mask), C1~(Diagonal), D0~(Fixed-N Mask), and D1~(Geo Mask) are the strategies applied to the object-object block of the mask.} \label{tab:alternatives}
\end{table*}

\subsection{Comparison with the State-of-the-Art}
\cref{tab:performance_comparison} provides a comprehensive comparison between our method and state-of-the-art approaches, including
(1) point-based models~\cite{3dllm,scenellm,llava3d}, which aggregate 2D image features in the point or voxel space;
(2) video-based models~\cite{gpt4scene,video3dllm,ross3d}, which treat multi-view images as video sequences; and
(3) object-based models~\cite{chat3d,chat3dv2,grounded3dllm,leo,inst3dlmm, robin3d, chatscene, 3dgraphllm}, which decompose 3D scenes into object-level representations.
Our approach is evaluated on two object-centric frameworks: Chat-Scene~\cite{chatscene} and 3DGraphLLM~\cite{3dgraphllm}.

Compared with video-based models, our model achieves higher accuracy on visual grounding tasks~(ScanRefer and Multi3DRefer), but shows lower performance on question-answering tasks~(ScanQA and SQA3D).
We conjecture that this performance gap arises from differences in the underlying language models.
Object-based approaches rely on LLMs~\cite{vicuna, llama3} trained primarily on text data, whereas video-based methods adopt MLLMs~\cite{qwen2-vl, llavavideo} further trained on large-scale image and video QA datasets, which equip them with stronger multi-modal reasoning capabilities.
Indeed, MLLMs already exhibit strong zero-shot results on ScanQA and SQA3D without any fine-tuning, as illustrated in the first two rows of the video-based section in \cref{tab:performance_comparison}.
As for object-based methods, our model consistently surpasses prior baselines, with particularly notable improvements in visual grounding.
These results underscore the importance of adapting the decoder’s attention mechanism to 3D reasoning, which proves highly effective for advancing 3D scene understanding.

\begin{table*}[t]
\centering
\small
\setlength{\tabcolsep}{4pt}
\renewcommand{\arraystretch}{0.9}
\begin{tabular}{cc|cccccccccc}
\toprule
    \multirow{2}{*}{Geo Mask} &
    \multirow{2}{*}{Inst Mask} &
    \multicolumn{2}{c}{ScanRefer$_{\text{val}}$} &
    \multicolumn{2}{c}{Multi3DRefer$_{\text{val}}$} &
    \multicolumn{2}{c}{Scan2Cap$_{\text{val}}$} &
    \multicolumn{2}{c}{ScanQA$_{\text{val}}$} &
    \multicolumn{2}{c}{SQA3D$_{\text{val}}$} \\
      &  & Acc@0.25 & Acc@0.5 & F1@0.25 & F1@0.5 & C@0.5 & B-4@0.5 & C & B-4 & EM   & EM-R \\ 
\midrule
 $\cdot$ & $\cdot$ & 55.3 & 49.5 & 59.6 & 54.8 & 78.1 & 35.6 & 88.3 & 14.1 & 53.7 & 56.4 \\
\checkmark & $\cdot$ & 58.6 & 53.1 & 62.0 & 57.3 & 82.4 & 37.0 & \textbf{94.2} & 15.0 & \textbf{55.9} & 58.6 \\
$\cdot$ & \checkmark & 57.6 & 51.8 & 62.0 & 57.0 & 81.1 & 36.4 & 91.1 & 14.9 & 55.1 & 58.0 \\
\checkmark & \checkmark  & \textbf{59.6} & \textbf{54.1} & \textbf{63.7} & \textbf{58.7} & \textbf{84.2} & \textbf{38.0} & 94.0 & \textbf{15.2} & \textbf{55.9} & \textbf{58.9} \\
\bottomrule
\end{tabular}
\caption{\textbf{Ablation study on 3D-SLIM.} To evaluate the contribution of each component, we conduct ablation studies on Geometry-adaptive Mask~(Geo Mask) and Instruction-aware Mask~(Inst Mask) within the Chat-Scene framework.}
\vspace{-0.3em}
\label{tab:ablation study chat-scene}
\end{table*}
\begin{table*}[t]
\label{tab:ablation study}
\centering
\small
\setlength{\tabcolsep}{4pt}
\renewcommand{\arraystretch}{0.9}
\begin{tabular}{cc|cccccccccc}
\toprule
    \multirow{2}{*}{Row} &
    \multirow{2}{*}{$k_{\min}$, $k_{\max}$} &
    \multicolumn{2}{c}{ScanRefer$_{\text{val}}$} &
    \multicolumn{2}{c}{Multi3DRefer$_{\text{val}}$} &
    \multicolumn{2}{c}{Scan2Cap$_{\text{val}}$} &
    \multicolumn{2}{c}{ScanQA$_{\text{val}}$} &
    \multicolumn{2}{c}{SQA3D$_{\text{val}}$} \\
      & &  Acc@0.25 & Acc@0.5 & F1@0.25 & F1@0.5 & C@0.5 & B-4@0.5 & C & B-4 & EM   & EM-R \\ 
      \midrule
      R0 & 0, 5 & 57.0 & 51.1 & 60.8 & 55.6 & 81.4 & 37.1 & 89.3 & 14.5 & 55.2 & 58.3 \\
      R1 & 0, 10 & 57.5 & 51.6 & 61.8 & 56.7 & 83.2 & 37.4 & 90.3 & 14.4 & 55.3 & 58.2 \\
      R2 & 2, 10 & 58.6 & 53.1 & 62.0 & 57.3 & 82.4 & 37.0 & 94.2 & 15.0 & 55.9 & 58.6 \\
      R3 & 2, 20 & 58.8 & 53.1 & 62.2 & 57.1 & 82.8 & 37.0 & 92.4 & 14.3 & 55.2 & 58.9 \\
\bottomrule
\end{tabular}
\caption{\textbf{Attention Range for Geo Mask.} 
Performance of the Geo Mask with different attentive neighborhood bounds~($k_{\text{min}}$, $k_{\text{max}}$) on the Chat-Scene baseline.
} 
\vspace{-0.3em}
\label{tab:hyperparam}
\end{table*}

\subsection{Ablation Study}
\paragraph{Effectiveness with different LLMs.}
\cref{tab:various llm} demonstrates that 3D-SLIM consistently boosts performance across diverse LLM backbones, including Vicuna-7B~\cite{vicuna}, Llama3-8B~\cite{llama3}, and Qwen2-7B~\cite{qwen2}.
When applied to Vicuna-7B, 3D-SLIM yields notable improvements across all benchmarks.
Similar performance gains are also observed with Llama3-8B and Qwen3-8B, demonstrating consistent enhancements across both grounding and reasoning tasks.
These results highlight that 3D-SLIM operates as a decoder-agnostic and broadly effective module for 3D scene understanding.

\paragraph{Analysis of Decoder Masking Strategies.}
To confirm the effectiveness of our Geometry-adaptive Mask~(Geo Mask), we compare several alternative designs in \cref{tab:alternatives}.
All experiments are conducted on the Chat-Scene baseline using Vicuna-7B as the decoder LLM.
Note that the Instruction-aware Mask~(Inst Mask) is excluded to isolate the influence of Geo Mask design choices.

\noindent\textbf{(1) Full Attention on All Tokens.}
This experiment investigates whether applying full attention to all tokens, without any dedicated mask, can recover the spatial and cross-modal dependencies lost due to the causal mask.
The standard causal mask~(A0) is compared with a full attention variant~(B0).
Although full attention removes artificial order and restores cross-modal interactions, empirical results show that it leads to performance drops in captioning and question-answering tasks~(Scan2Cap, ScanQA, and SQA3D).
This indicates that fully unmasked attention fails to provide useful structural cues and can even hinder decoder reasoning.

\noindent\textbf{(2) Removing the Object Order Constraint.}
The next analysis focuses on whether relaxing the causal constraint for object-object interactions helps the decoder exploit object-level relations.
This is motivated by the assumption that the causal mask is essential for non-object tokens~(e.g., system and text prompts) but may overly restrict object-centric reasoning.
Based on this assumption, the object-object attention block is examined through two variants: C0~(Full Mask), which enables all-to-all interactions, and C1~(Diagonal Mask), which blocks all inter-object attention.
The results confirm that removing the causal constraint from this block is beneficial. 
However, they also show an interesting observation: the Full Mask performs almost identically to the Diagonal Mask. 
This suggests that while removing the sequential constraint is a step in the right direction, merely expanding the attention scope to be all-to-all is insufficient to capture the structural cues essential for spatial reasoning.

\noindent\textbf{(3) Modeling Local Spatial Relationships.}
The preceding analyses reveal the limitations of na\"ive mask modifications, underscoring the need for explicit locality guidance.
To explore this, we compare two locality-based masking strategies:
D0~(Fixed-N Mask), which connects each object to its predefined top-5 nearest neighbors, and D1~(Geo Mask), which adapts the attentive neighborhood according to each object's local spatial density.
As shown in \cref{tab:alternatives}, both variants surpass the C0~(Full Mask) and C1~(Diagonal Mask), demonstrating that explicit locality guidance improves spatial structure modeling.
Moreover, Geo Mask consistently outperforms Fixed-N Mask, highlighting that adjusting the attention range based on local density provides a more accurate understanding of non-uniform 3D structures and strengthens spatial reasoning in complex scenes.

\paragraph{Component-level Analysis.}
To validate the contribution of each component in 3D-SLIM, we conduct ablation studies on Chat-Scene baseline.
As reported in \cref{tab:ablation study chat-scene}, both Geo Mask and Inst Mask consistently improve overall performance.
Moreover, when applied together, they yield additional gains, reaching state-of-the-art results on visual grounding benchmarks.
This complementary behavior suggests that the two masking schemes address distinct yet cooperative aspects of 3D scene decoding, emphasizing the need to model both the spatial structure and instruction awareness in 3D scene understanding.

\paragraph{Attention Range for Geo Mask.}
\cref{tab:hyperparam} presents an ablation of the Geo Mask's attentive neighborhood bounds~($k_{\min}, k_{\max}$) on the Chat-Scene baseline.
First, comparing R0~($k_{\min}=0$, $k_{\max}=5$) and R1~(0, 10) shows that overly narrow bounds limit information exchange between nearby objects, whereas larger ranges enable richer spatial interactions and improve performance.
Furthermore, R2~(2,10) reveals that ensuring a minimum neighborhood~($k_{\min}$) stabilizes attention, especially for the object in sparse regions, leading to consistent gains.
However, R3~(2, 20) indicates that excessively large upper bounds~($k_{\max}$) cause attention to spread too widely, resulting in degraded performance.
Overall, the range of [2, 10] offers the most balanced performance across all benchmarks, and we adopt these values as our default configuration.

\begin{figure}[t]
    \centering
    \includegraphics[width=1.0\linewidth]{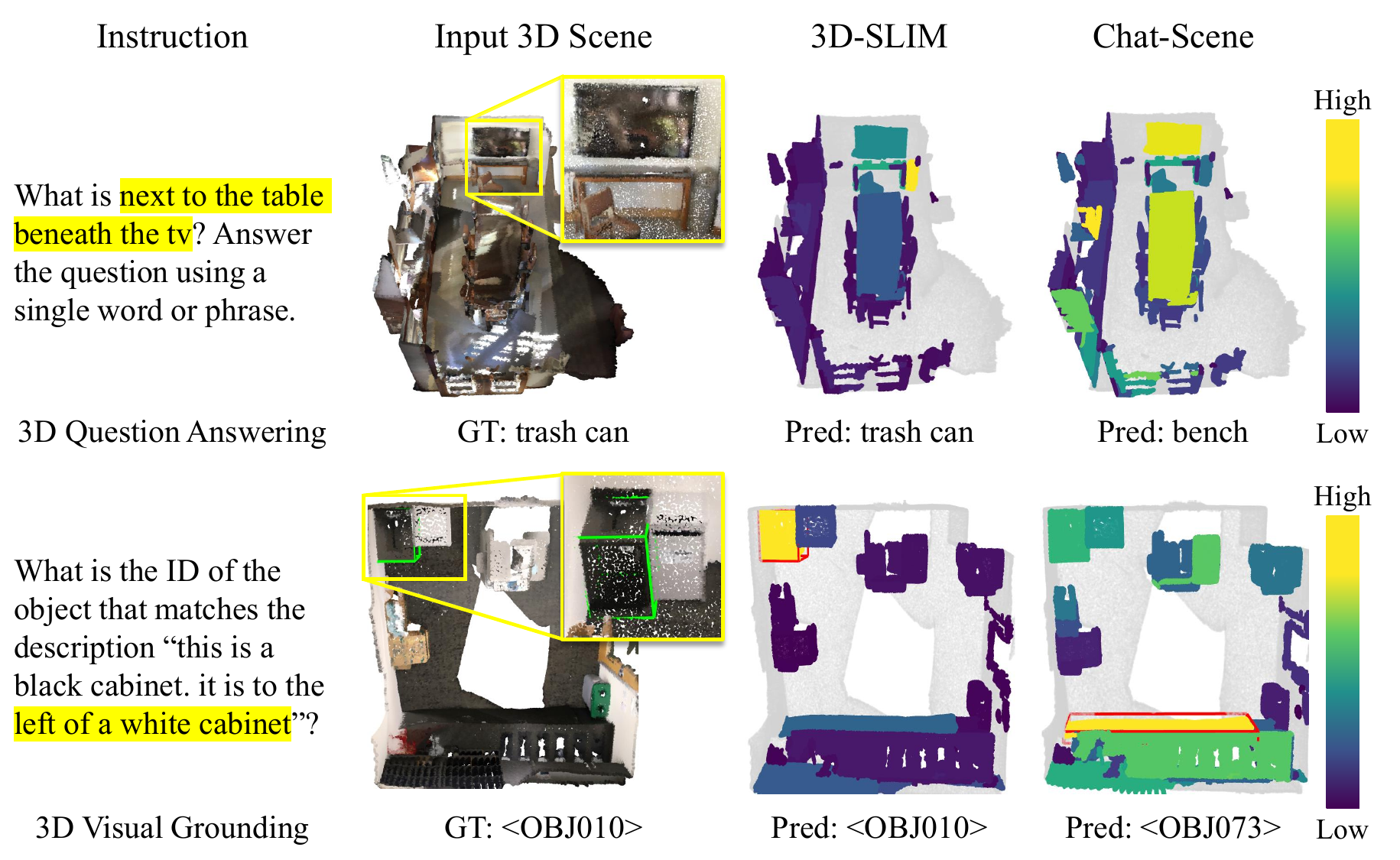}
    \caption{
    \textbf{Visualization of LLM attention map.} 
    In the instruction, key cues for answering the question are highlighted in yellow. The attention map is colored by activation intensity, where yellow represents high values and purple represents low ones. The green box in the second row denote the GT object, whereas the red ones indicate the predicted objects.
    }
    \label{fig:qualitative}
\end{figure}
\paragraph{Qualitative Results.}
To examine how the model allocates attention to perform its tasks, we visualize the attention maps from the output tokens to the 3D object tokens, computed by averaging attention weights across all layers and heads, as shown in \cref{fig:qualitative}.
The first row shows that 3D-SLIM accurately interprets the complex spatial phrase "next to the table beneath the tv" for the question-answering task.
3D-SLIM precisely focuses on the trash can and its surrounding region, which is consistent with the instructions. 
In contrast, Chat-Scene mainly attends to salient nouns such as "table" and "tv", missing the contextual meaning and producing an incorrect answer.
For the second example, the visual grounding task, the scene includes two black cabinets, where the model must identify the one "to the left of a white cabinet". 
3D-SLIM correctly focuses on the described region and grounds the target object by capturing the relational context between the two cabinets.
In contrast, Chat-Scene attends to a black cabinet in isolation, failing to encode its spatial relationship to the white cabinet and thus selecting the wrong object.
\section{Conclusion}
In this work, we identified a fundamental conflict in existing 3D scene-language frameworks: the standard LLM decoder imposes spurious sequential dependency on inherently order-agnostic 3D objects and blocks object-instruction interactions crucial for task-specific reasoning.
Motivated by this limitation, we proposed 3D-SLIM, a simple, parameter-free masking strategy. 
3D-SLIM replaces the standard causal mask with two specialized components: (1) Geometry-adaptive Mask~(Geo Mask) that models object-object interactions based on spatial proximity and local density, and (2) Instruction-aware Mask~(Inst Mask) that enables object tokens to directly attend to instruction tokens, facilitating task-guided scene representation.
Extensive experiments demonstrated that 3D-SLIM consistently improves object-centric frameworks, highlighting the importance of decoder design in 3D-LLMs. 
We believe 3D-SLIM provides strong foundation for future 3D multi-modal models, paving the way for more capable and efficient spatial reasoning in applications like embodied AI and robotics.
\section*{Acknowledgements}
This work was supported in part by MSIT/IITP (No. RS-2022-II220680, RS-2020-II201821, RS-2019-II190421, RS-2024-00459618, RS-2024-00360227, RS-2024-00437633, RS-2024-00437102, RS-2025-25442569), MSIT/NRF (No. RS-2024-00357729), and KNPA/KIPoT (No. RS-2025-25393280).
{
    \small
    \bibliographystyle{ieeenat_fullname}
    \bibliography{main}
}

\clearpage
\setcounter{page}{1}
\maketitlesupplementary
\begin{figure*}[t]
    \centering
    \includegraphics[width=1.0\linewidth]{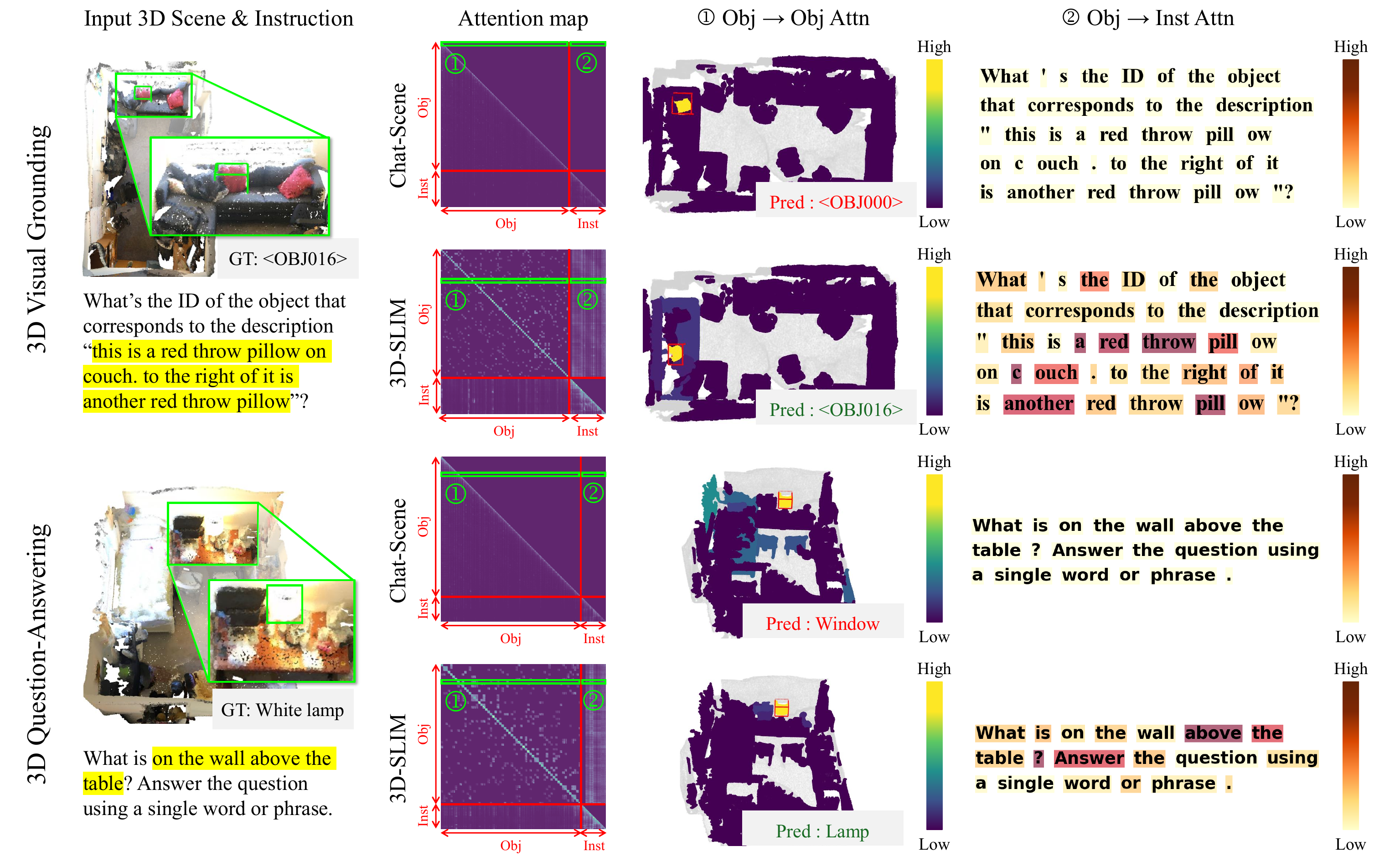}
    \caption{
    \textbf{Attention map over 3D objects and instruction.}
    For simplicity, the attention map is computed over all object and instruction tokens except the system tokens, where yellow indicates high values and purple indicates low ones. The green boxes labeled ① and ② correspond to the regions used for Obj$\rightarrow$Obj Attn and Obj$\rightarrow$Inst Attn.
    Obj$\rightarrow$Obj Attn displays where the reference object token focuses within the 3D scene. For visual grounding, the reference is the model-predicted object; for question answering, it is a manually identified target object. Gray points indicate background.
    Obj$\rightarrow$Inst Attn visualizes how the reference  token attends to instruction tokens, with red and yellow indicating high and low activations.
    }
    \label{fig:supple_qualitative}
\end{figure*}
\section{Qualitative Analysis}
We further analyze how the model allocates attention by visualizing attention maps averaged over all layers and heads, as shown in \cref{fig:supple_qualitative}.
This analysis allows us to (i) inspect activation patterns between 3D objects and instruction tokens, (ii) examine how a reference object token attends to other objects to encode spatial relations, and (iii) observe how the reference token interacts with instruction tokens to incorporate textual cues.

\paragraph{3D Visual Grounding.}
In the first example, the scene contains two red throw pillows, and the model must identify the one described by “to the right of it is another red throw pillow”. 
3D-SLIM correctly grounds the target as <OBJ016>, whereas Chat-Scene incorrectly selects <OBJ000>.
To understand this difference, we examine their attention patterns.

Chat-Scene adopts a causal mask that prevents each object token from attending to later tokens in the sequence.
Since <OBJ000> is the first object token, the causal mask forces its Obj$\rightarrow$Obj attention to all other object tokens to be zero.
As a result, the token relies solely on itself and fails to capture the spatial relation between the two pillows.
The causal mask also suppresses Obj$\rightarrow$Inst attention, limiting access to instruction tokens and weakening the effect of textual cues.

In contrast, 3D-SLIM combines Geo Mask, which injects spatial proximity into object-object attention, with Inst Mask, which explicitly permits object-instruction interaction.
The Obj$\rightarrow$Obj attention of <OBJ016> is concentrated around the target region, reflecting the underlying 3D structure.
Meanwhile, its Obj$\rightarrow$Inst attention strongly activates keywords such as “red", "throw", "pillow", “couch" and “another", showing that the model effectively incorporates the instruction.
Overall, this qualitative comparison indicates that 3D-SLIM mitigates the spurious order-dependent correlations introduced by the causal mask and restores informative interactions between instruction and object tokens, leading to more reliable performance on the grounding task.

\paragraph{3D Question-Answering.}
In the second example, the correct answer to the question is "White lamp".
We manually identify the corresponding object token~(<OBJ006>) and visualize its Obj$\rightarrow$Obj and Obj$\rightarrow$Inst attention. 

For Obj$\rightarrow$Obj attention, Chat-Scene restricts the lamp token to attending only to objects preceding it in the sequence.
This constraint prevents the token from focusing on spatially relevant neighbors and instead shifts attention toward distant regions.
By contrast, 3D-SLIM associates the lamp token with nearby objects, aligning its attention pattern with the actual 3D arrangement rather than the token order.

For Obj$\rightarrow$Inst attention, Chat-Scene assigns zero weight to all instruction tokens, so the lamp token is encoded without any guidance from the instruction.
In comparison, 3D-SLIM places high attention on key phrases such as “above” and “table”, indicating that instruction cues are directly reflected in the lamp token representation.
Taken together, these observations highlight that 3D-SLIM leverages both spatial structure among object tokens and semantic guidance from the instruction, which enables the model to infer the correct answer.

\end{document}